\documentclass[default,iicol]{sn-jnl}% Default with double column layout

\usepackage[justification=centering]{caption} 
\usepackage{amsfonts}
\usepackage{subfigure}
\usepackage{multirow}
\usepackage{booktabs}
\usepackage{epstopdf}
\AppendGraphicsExtensions{.jpg}
\usepackage{graphicx}%
\usepackage{multirow}%
\usepackage{amsmath,amssymb,amsfonts}%
\usepackage{amsthm}%
\usepackage{mathrsfs}%
\usepackage[title]{appendix}%
\usepackage{xcolor}%
\usepackage{textcomp}%
\usepackage{manyfoot}%
\usepackage{booktabs}%
\usepackage{algorithm}%
\usepackage{algorithmicx}%
\usepackage{algpseudocode}%
\usepackage{listings}%

\usepackage{subfigure}
\usepackage{multirow}
\usepackage{booktabs}
\usepackage{caption}
\DeclareCaptionLabelSeparator{twospace}{\ ~}   
\usepackage{hyperref}
\usepackage[hyphenbreaks]{breakurl}
\theoremstyle{thmstyleone}%
\theoremstyle{thmstyletwo}%

\theoremstyle{thmstylethree}%

\begin{document}

\title[Article Title]{Few-Shot Medical Image Segmentation with Large Kernel Attention}

%%=============================================================%%
%% Prefix	-> \pfx{Dr}
%% GivenName	-> \fnm{Joergen W.}
%% Particle	-> \spfx{van der} -> surname prefix
%% FamilyName	-> \sur{Ploeg}
%% Suffix	-> \sfx{IV}
%% NatureName	-> \tanm{Poet Laureate} -> Title after name
%% Degrees	-> \dgr{MSc, PhD}
%% \author*[1,2]{\pfx{Dr} \fnm{Joergen W.} \spfx{van der} \sur{Ploeg} \sfx{IV} \tanm{Poet Laureate} 
%%                 \dgr{MSc, PhD}}\email{iauthor@gmail.com}
%%=============================================================%%

\author[1]{\fnm{Xiaoxiao} \sur{Wu}}\email{wxx\_52v@126.com}

\author[1]{\fnm{Xiaowei} \sur{Chen}}\email{932197321@qq.com}

\author*[1]{\fnm{Zhenguo} \sur{Gao}}\email{gzg@hqu.edu.cn}

\author[1]{\fnm{Shulei} \sur{Qu}}\email{13562506757@163.com}

\author[1]{\fnm{Yuanyuan} \sur{Qiu}}\email{1359379059@qq.com}

\affil*[1]{\orgdiv{College of Computer Science and Technology}, \orgname{Huaqiao University}, \orgaddress{\street{No.668 Jimei Avenue}, \city{Xiamen}, \postcode{361021}, \state{Fujian}, \country{China}}}

% \affil[2]{\orgdiv{Department}, \orgname{Organization}, \orgaddress{\street{Street}, \city{City}, \postcode{10587}, \state{State}, \country{Country}}}

% \affil[3]{\orgdiv{Department}, \orgname{Organization}, \orgaddress{\street{Street}, \city{City}, \postcode{610101}, \state{State}, \country{Country}}}

%%==================================%%
%% sample for unstructured abstract %%
%%==================================%%

\abstract{Medical image segmentation has witnessed significant advancements with the emergence of deep learning. However, the reliance of most neural network models on a substantial amount of annotated data remains a challenge for medical image segmentation. To address this issue, few-shot segmentation methods based on meta-learning have been employed. Presently, the methods primarily focus on aligning the support set and query set to enhance performance, but this approach hinders further improvement of the model’s effectiveness. In this paper, our objective is to propose a few-shot medical segmentation model that acquire comprehensive feature representation capabilities, which will boost segmentation accuracy by capturing both local and long-range features. To achieve this, we introduce a plug-and-play attention module that dynamically enhances both query and support features, thereby improving the representativeness of the extracted features. Our model comprises four key modules: a dual-path feature extractor, an attention module, an adaptive prototype prediction module, and a multi-scale prediction fusion module. Specifically, the dual-path feature extractor acquires multi-scale features by obtaining features of 32*32 size and 64*64 size. The attention module follows the feature extractor and captures local and long-range information. The adaptive prototype prediction module automatically adjusts the anomaly score threshold to predict prototypes, while the multi-scale fusion prediction module integrates prediction masks of various scales to produce the final segmentation result. We conducted experiments on publicly available MRI datasets, namely CHAOS and CMR, and compared our method with other advanced techniques. The results demonstrate that our method achieves state-of-the-art performance.}


%%\pacs[JEL Classification]{D8, H51}

%%\pacs[MSC Classification]{35A01, 65L10, 65L12, 65L20, 65L70}

\maketitle
\clearpage
\section{Introduction}

Convolutional neural networks (CNNs) have been widely used in deep learning for a variety of image segmentation tasks, particularly within the medical field. Medical research, clinical diagnosis, and case analysis all benefit greatly from the use of medical image segmentation. Quantitative measurement and analysis of relevant imaging indicators before and after therapy are crucial for medical diagnosis and patient treatment plan revision. Although CNNs have proven successful for segmentation tasks\cite{b1,b2,b3,b4}, most current models rely heavily on annotated data. However, in medical filed, data labeling often requires manual work by medical experts, making it expensive and time-consuming\cite{b5,b6}. Therefore, scholars have been exploring ways to complete the image segmentation task with limited amount of annotation data.

Few-shot learning(FSS), also known as learn to learn, is a way to learn from a few annotated examples and generalize to new classes. Most models of few-shot segmentation are constructed by meta-learning techniques (Meta-FSS) \cite{b7,b8,b9,b10,b11,b12}. In the meta-training stage, the model can learn task-independent generation ability by constructing different meta-tasks. Each task (also known as a scenario) has a support set and a query set. After learning the representation of each class on the annotated support set, the query set is segmented using the previously acquired information. With this paradigm, it is possible to segment a previous “unseen” organ with a few annotated images in the test stage. However, in order to avoid overfitting, the Meta-FSS methods rely on a large number of annotated data, which is unrealistic in the medical field.

To solve this problem, Ouyang et al. \cite{b7} creatively proposed a self-supervised training approach to bypass model’s need for large amounts of labeled data. Specifically, they created image pseudo labels using superpixels and trained with those labels. Superpixel is a type of short and compact image fragment, and the pseudo labels produced by superpixel typically contain real semantic information, which is helpful to the training of the model. Based on Ouyang's work, Hansen et al.\cite{b8} further proposed the concept of supervoxel. By using supervoxel to generate pseudo labels, local information and volume information of the images can be both used. Supervoxel segmentation has gradually gained popularity in the medical image segmentation in recent years\cite{b13,b14,b15}. Their proposed approaches effectively address the issue of data annotation, yet the challenge of data scarcity persists. 

There are two major factors that hinder model performance: intra-class and inter-class gaps\cite{b16}.The former arises from insufficient data in the support set, resulting in disparities between the support set and query set. The latter stems from non-overlapping categories between the training and test sets. To minimum these gaps, previous studies have introduced attention mechanisms\cite{b17,b18,b19,b32}, while others have adopted multi-scale methods\cite{b9,b20,b21}. Prior work has utilized attention mechanisms to learn common features between query sets and support sets for reducing intra-class gaps. Although this method has shown some effectiveness, it struggles to adapt to complex and variable medical sections encountered in clinical settings where even identical organ slices can vary significantly. Moreover, scarce medical data can bias model focus due to subtle differences between support and query images, particularly in 1-shot scenarios. In this study, we propose a multi-scale approach based on the large kernel attention(LKA) mechanism by combining these two methods. Different from previous approaches, our method employs attention mechanisms to extract more comprehensive features, allowing the model itself to determine their importance. Additionally, we emphasize the importance of constructing a high-quality foreground in medical image segmentation, given the pronounced foreground-background imbalance. By integrating multi-scale information, local information, and long-range information, our method achieves a comprehensive feature representation, thereby attaining state-of-the-art performance.

\begin{figure}[t]
\setlength{\belowcaptionskip}{5mm}
\centerline{\includegraphics[width=\columnwidth]{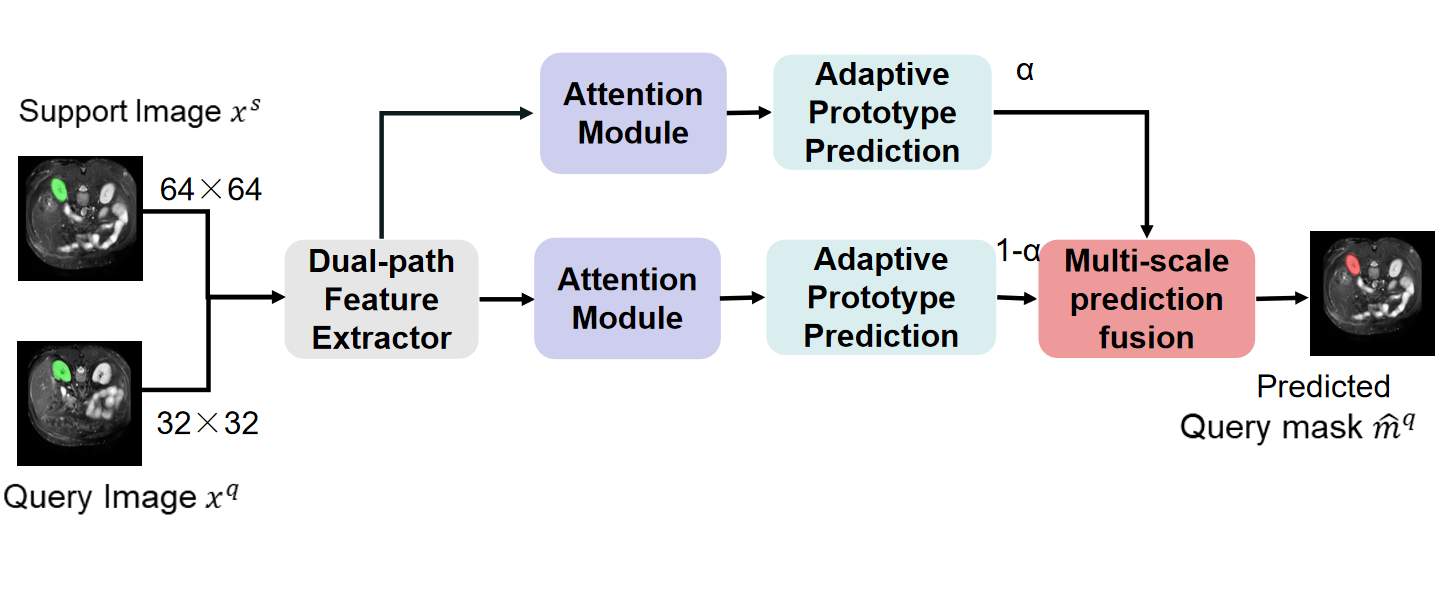}}
\captionsetup{justification=raggedright}
\caption{The model we proposed.}
\label{fig1}
\end{figure}

In summary, we propose a multi-scale model based on the large kernel attention mechanism, as shown in Figure \ref{fig1}. The model consists of four modules: dual-path feature extractor, attention module, adaptive prototype prediction module and multi-scale prediction fusion module. Among them, the attention module follows the feature extractor, which can not only enhance the semantic representation of feature vectors, but also capture local features and long-range semantic features. The adaptive prototype prediction module can predict the prototype by automatically adjusting the threshold of anomaly scores, and the multi-scale fusion prediction module can combine the prediction masks of different scales to get the final prediction segmentation result. We test the model on the public datasets.

The main contributions of this paper are as follows:

(1)We propose a large kernel attention module to enhance support features and query features, which can capture both local features and long-range semantic features.

(2)By integrating the large kernel attention mechanism with multi-scale analysis, our approach enables the extraction of features encompassing local and long-range information as well as multi-scale information, thereby achieving a comprehensive representation that surpasses existing methods.

(3)We propose an effective few-shot segmentation model and obtain state-of-the-art performance on two widely used organ segmentation datasets.

 \renewcommand{\dblfloatpagefraction}{.9}

\section{Related Work}
We first review the development of convolutional neural networks in the medical field. Then we focus on the few-shot learning and introduce the recent development of few-shot medical segmentation model. We also introduce the application of attention mechanisms in medical image segmentation, focusing on more widely used attention methods. Finally, we introduce the segmentation method based on supervoxels and its advantages briefly.
\subsection{Medical Image Segmentation}
In recent years, deep learning technology has attracted more and more researchers’attention. In the medical field, it is common to use convolutional neural networks to perform segmentation tasks. Full convolutional network (FCN) \cite{b3} is a kind of network with strong learning ability, and it is also the first network to realize end-to-end pixel level segmentation. U-Net, proposed by Ronneberger et al.\cite{b22}, uses a symmetrical encoder-decoder structure. In the contraction path, features are extracted using convolution. In the expansion path, the expression ability of the feature map is enhanced by upsampling the feature map and concatenating it with the corresponding feature map from the contraction path. The final feature map is resized to the original image size for segmentation. U-Net has inspired further research in this field. U-Net’s variants include Res-Unet\cite{b25}, Dense-Unet\cite{b26} and MultiRes-UNet\cite{b27} et al. Çiçek et al.\cite{b23} propose 3D U-Net, replacing all 2-dimensional operations in the original U-Net with 3-dimensional operations to achieve volume segmentation. Milletari et al.\cite{b24} propose the V-net structure. In addition, they propose a new objective function to deal with the severe imbalance between foreground and background voxels.

\subsection{Few-shot Semantic Segmentation}
Few-shot semantic segmentation extends few-shot learning from classification task to segmentation task. Specifically, given a set of labeled data, the task of few-shot segmentation is to segment objects on a new class. In some recent studies, Wang et al.\cite{b10} propose a prototype alignment network (PANet) from the perspective of metric learning to better utilize the information of support sets. Based on PANet, Ouyang\cite{b7} and Hansen\cite{b8} propose segmentation models based on superpixels and supervoxels by using self-supervised meta-learning method. Among them, Hansen et al.\cite{b8} believe that since the foreground types of medical images (such as organs) are relatively homogeneous, the imbalance between foreground and background is prominent. Therefore, they can avoid modeling the background and only build a prototype for the same structure of the foreground class. Specifically, only one class prototype is built in each ”scenario” and an anomaly score is introduced to measure the difference between the foreground prototype and the query feature. Shen et al.\cite{b9}, inspired by the learning mechanism of clinical experts, believes that the class prototype should be dynamically adjusted during the test stage to adapt to the differences between the training set and the test set. Protonotarios et al.\cite{b11} use the UNet architecture to solve lung cancer lesion segmentation problems, the difference is that it can continuously adjust the weights of the model based on user feedback, aiming to achieve the best performance. Li et al.\cite{b12}  utilize the feature pyramid technique, the sensitivity of the multi-scale feature space is enhanced for better performance in segmenting lung cancer lesions. The method proposed by Ding et al.\cite{b32} is modified on the basis of non-local block. By intensively comparing the similarity between query pixels and support pixels, the query features and support features are selectively enhanced.

\subsection{Attention Mechanisms}
Attention mechanism has been widely used since it was proposed. In a nutshell, the main idea of the attention mechanism is to highlight important features and ignore irrelevant ones. Recently, some studies have proposed attentional mechanisms that can enhance the network’s focus on segmented objects at little cost. For example, the CBAM module proposed by Woo et al.\cite{b17} can infer the attention maps along the two dimensions of channel and space, and then carry out adaptive feature fusion. Due to its low cost, it can be easily embedded into other networks to achieve end-to-end training. Attention Gate is first used in natural image analysis, knowledge graph and classification tasks. Roy et al.\cite{b18} use dense connections with squeeze and excitation blocks to make the interaction between the two branches stronger and achieve better performance. Sinha et al.\cite{b20} believe that the multi-scale method applied in the previous model is not effective because of the redundancy of information caused by the constant extraction of similar low-level features. However, the application of attention mechanism can well highlight the features of the segmented area and suppress other irrelevant parts. Therefore, the author uses a relatively perfect attention mechanism to make the model learn richer context information.

\subsection{Supervoxel Segmentation}
Both superpixels and supervoxels are collections of local pixels and voxels in an image. Specifically, a superpixel is a small, compact fragment of an image, usually obtained by clustering a local pixel by applying a statistical model. The supervoxel is the subvolume in the 3D image, which represents the similar voxel group in the local area of the image volume. The advantage of three-dimensional voxel is that it can make better use of the volumetric properties of medical images. In recent years, the use of supervoxel segmentation has gradually become a common method in the field of medical image segmentation\cite{b13,b14,b15}.

\section{Methods}
In this work, we propose a prototypical network based on large kernel attention mechanisms.

\begin{figure*}[!t]
\setlength{\belowcaptionskip}{5mm}
\centerline{\includegraphics[width=\linewidth]{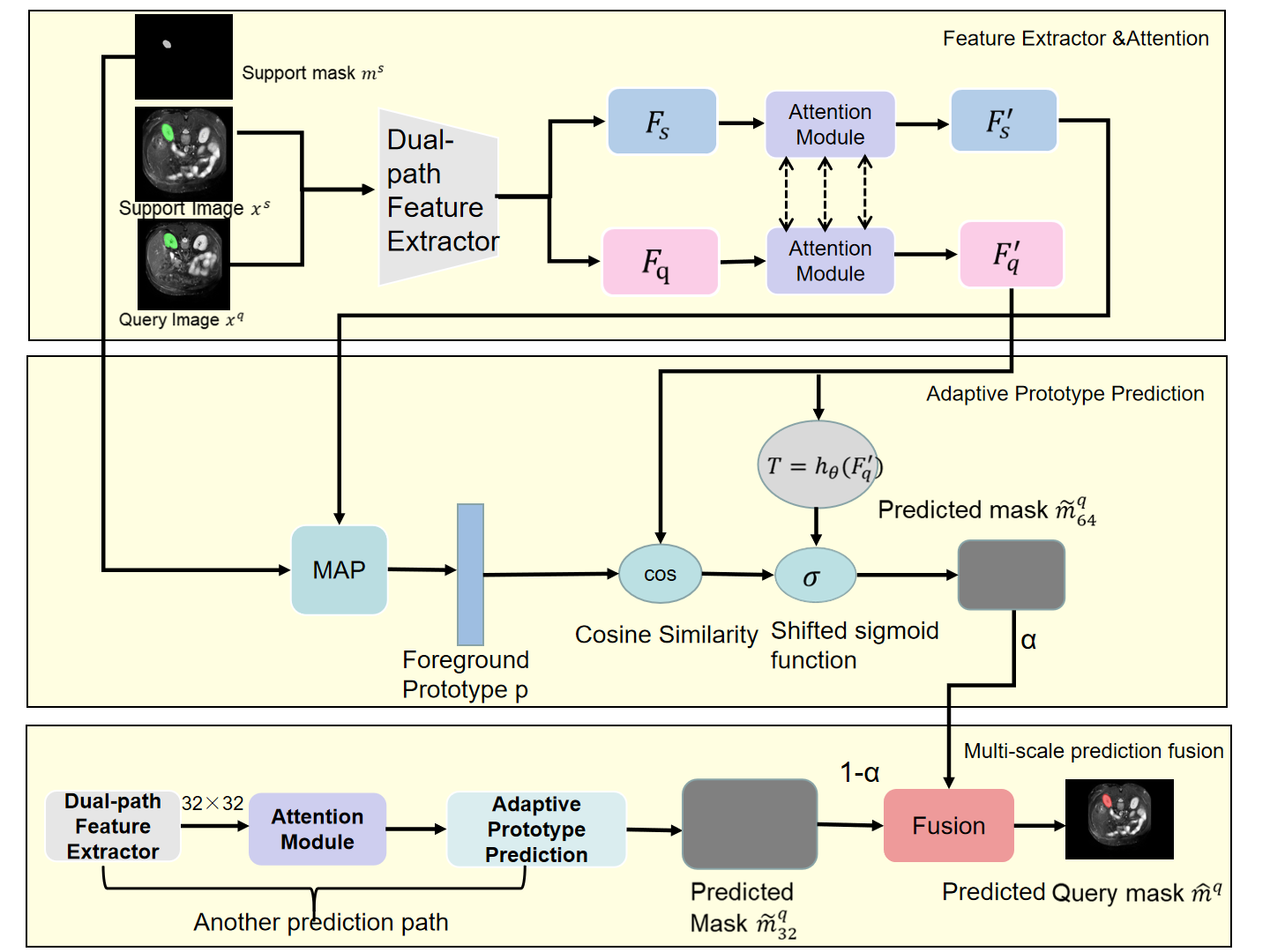}}
\captionsetup{justification=raggedright}
\caption{Illustration of the model. The model integrates feature information from two scales, 64*64 and 32*32. The feature flow of the 64*64 scale is illustrated here. The model consists of three layers: feature extractor and attention layer, adaptive prototype prediction layer, and multi-scale prediction fusion layer. The feature extractor and attention module extract comprehensive feature representations of query and support images. The adaptive prototype prediction layer generates predicted segmentation masks by calculating cosine similarity between foreground prototype and query feature and obtaining an adaptive threshold T. The multi-scale prediction fusion layer upsamples masks of different scales and fuses them to produce the final mask prediction result.}
\label{fig2}
\end{figure*}

The model consists of these main parts: (1) Dual-path feature extractor, which extracts features of 64*64 and 32*32 scale for both support images and query images; (2) attention module, which captures both local and long-range characteristics; (3) Adaptive prototype prediction layer, which uses class prototypes and adaptive anomaly scores to separate foreground and background; and (4) multi-scale prediction fusion layer, which fuses prediction masks of different scales for final segmentation results. See Figure \ref{fig2} for a detailed illustration of the model.

\subsection{Problem Definition}
% In few-shot segmentation, given a query set Q and a support set S, each with their corresponding binary segmentation masks, the task is to predict a segmentation mask for the query image in the query set. 

In few-shot segmentation, a segmentation model is initially trained on the training dataset $D_{{train}}$ and subsequently evaluated on the test dataset $D_{{test}}$ in episodes. The training dataset $D_{{train}}$ comprises training classes $C_{{train}}$ , while the test dataset $D_{{test}}$ includes test classes $C_{{test}}$ . It's worth noting that the training classes ${C_{{train}}}$  and the test classes ${C_{{test}}}$ are disjoint, denoted as ${C_{train}}\cap{C_{test}}$ = $\emptyset$. 

Each episode consists of a support set $S{}_i = \{ (x_1^j,m_1^j),(x_2^j,m_2^j), \ldots ,(x_k^j,m_k^j)\}$ and a query set $Q = \{ {x_q^j},{m_q^j}\}$ with the same class ${c_j} \in {C_{train}},j = 1,2, \ldots ,N$. Let image $X = \{ x_1^j,x_2^j, ... ,x_k^j,x_q^j, x_s, x_q\}$, $x \in X$, and mask $M = \{ m_1^j,m_2^j, ..., m_k^j,m_q^j,m_s,m_q\}$, $m \in M$, where the image ${x} \in \mathbb{R}{^{H \times W \times 1}}$ is a single-channel grayscale image, and the binary mask of the corresponding category is $m \in {\{ 0,1\} ^{H \times W}}$. Since each support set ${S_{i}}$ contains K annotated images and the number of object classes being segmented in query set Q is N, this is also called N-way K-shot. In general, the model first learns semantic information about each category from the support set and then uses the learned knowledge to segment the query set. The query mask is invisible to the model and will only be used to evaluate the model in the test stage. Specifically for our experiment, we adopt 1-way 1-shot meta-learning strategy.

% \subsection{Model structure}
% In this work, we propose a prototypical network based on attention mechanisms.

% The model consists of these main parts: (1) Feature extractor and attention module, which extract and combine features from query and support images while capturing both local and long-range characteristics; (2) Adaptive prototype prediction layer, which uses class prototypes and adaptive anomaly scores to separate foreground and background; and (3) multi-scale prediction fusion layer, which fuses prediction masks of different scales for final segmentation results. See Figure \ref{fig2} for a detailed illustration of the model.

\subsection{Dual-path Feature Extractor}
We employ a dual-scale feature extractor ${f_\theta }$ to extract features at both 64*64 and 32*32 scales from both support and query images. The extracted support image features ${F_s} = {f_\theta }({x^s})$, ${F_s} \in  \mathbb{R}{^{H' \times W' \times Z}}$, ${H'}$ represents the height of the extracted feature map, ${W'}$ represents the width of the extracted feature map, and Z represents the number of channels of the feature map. Corresponding, query image features ${F_q} = {f_\theta }({x^q})$. We extracted 64*64 (1/4 the size of the original picture) and 32*32 (1/8 the size of the original image) feature maps from the Resnet-101 pre-trained on Coco dataset network, respectively, then fused them to produce multi-scale features.

\begin{figure}[t]
\setlength{\belowcaptionskip}{5mm}
\centerline{\includegraphics[width=\columnwidth]{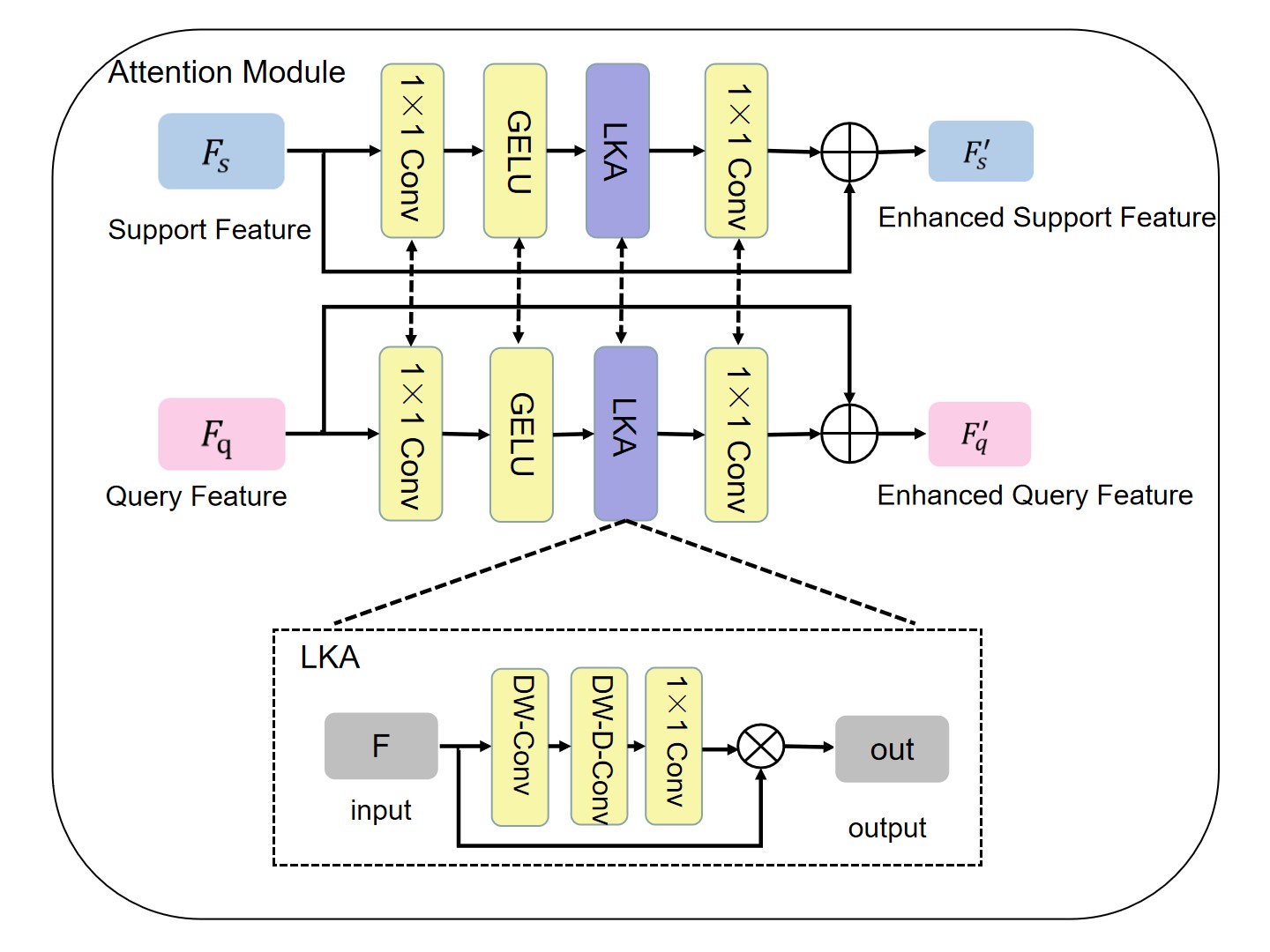}}
\captionsetup{justification=raggedright}
\caption{Attention module. Where, DW-Conv represents the depth-wise convolution, DW-D-Conv represents the depth-wise-dilation convolution, and 1×1 Conv represents the 1*1 convolution.}
\label{fig3}
\end{figure}

\subsection{Attention Module}
We aim to extract more comprehensive semantic features by integrating local and long-range features in our dual-path feature extractor. Inspired by the work of Park et al.\cite{b28} and Guo et al.\cite{b29}, we use attention methods at early layers to focus on local information and decompose large convolution kernels to capture long-range relationships. This results in enhanced support and query features, noted as ${F_s^{'}}$ and ${{F_q}^{'}}$ respectively. ${F_s^{'}}$ and ${{F_q}^{'}}$ are obtained by subjecting support and query features to a 1*1 convolution followed by a GELU activation function. This activation function has the form:

% \begin{equation}
% xP(X \le x) = x \cdot \phi (x) \label{eq}
% \end{equation}

% Where, $\phi (x)$ refers to the cumulative distribution of the Gaussian normal distribution of x, and the full form is shown below:

% \begin{equation}
% xP(X \le x) = x\int_{ - \infty }^{\rm{x}} {\frac{{{e^{ - \frac{{{{(X - \mu )}^2}}}{{2{\sigma ^2}}}}}}}{{\sqrt {2\pi } \sigma }}} dX \label{eq} 
% \end{equation}
\begin{equation}
    \text{GELU}(x) = \frac{1}{2} \cdot x \left(1 + \text{erf}\left(\frac{x}{\sqrt{2}}\right)\right) 
\end{equation}
The error function can be expressed as: 
\begin{equation}
    \text{erf}(x) = \frac{2}{\sqrt{\pi}} \int_{0}^{x} e^{-t^2} dt 
\end{equation}

The article describes a method to capture local, long-range, and channel features using a large kernel attention (LKA) module comprising a depth-wise convolution, a depth-wise-dilation convolution, and a 1*1 general convolution. Unlike traditional convolution methods, each input layer channel is independently convolved using depth-wise convolution, while depth-wise-dilation expands the receptive field. The 1*1 convolution is used for self-adaptability of channel dimension. The LKA mechanism combines local, long-range, and channel information to create a comprehensive feature representation.\cite{b29} It can be expressed as:
\begin{small}
\begin{equation}
LKA(F) = Conv{_{1 \times 1}}(\rm DW-D-Conv(\rm DW-Conv(F)))
% LKA(F) = Con{v_{1\times1}}({\rm{DW-D-Conv}}({\rm{DW-Conv}}(F)))
\end{equation}
\end{small}
\begin{equation}
Output = LKA \otimes F\label{eq}
\end{equation}

Where, $F \in \mathbb{R} {^{C \times H \times W}}$ represents the input features, $LKA(F) \in \mathbb{R} {^{C \times H \times W}}$ represents the attention feature map of $F$. $\otimes $ represents element-level multiplication. After element-level multiplication between attention feature map and support features or query features, the enhanced support features ${F_s^{'}}$ and query features ${F_q^{'}}$ are obtained. By using this attention module, not only the spatial dimension but also the channel dimension can be self-adaptive. Finally, the enhanced feature vectors are spliced into the prototype segmentation module. The attention module is shown in Figure \ref{fig3}.

\subsection{Adaptive Prototype Prediction Module}

To deal with the high homogeneity of medical images, a single prototype is constructed in each episode, as seen in Hansen\cite{b8} and Shen et al\cite{b9}. Metric learning techniques are employed to learn prototype and segment the image. The prototype for a class is calculated using masked averaging pooling between support feature ${F_s^{'}}$ and support masks ${m_s}$:

\begin{equation}
{\rm{p}} = \frac{{\sum {_{x,y}F_{\rm{s}}^{'}(x,y)}  \odot {m_s}(x,y)}}{{\sum {_{x,y}{m_s}(x,y)} }} \label{eq}
\end{equation}

Where ${F_s^{'}}$ is scaled to the mask ${m_s}$ size (H, W), and $F_{\rm{s}}^{'}(x,y)$ represents the pixels in the support feature ${F_s^{'}}$. And  ${\odot}$ represents the Hadamard product.

Then, we use the obtained class prototype to segment the query features. By computing the negative cosine similarity between the class prototype p and the query feature ${F_q^{'}}$, we created the anomaly score S:
\begin{equation}
{\rm{S(x,y) =   - }}\alpha \frac{{F_q^{'}(x,y) \cdot p}}{{\left\| {F_q^{'}(x,y)} \right\|\left\| p \right\|}} \label{eq}
\end{equation}

Where ${F_q^{'}(x,y)}$ represents pixels in the query features, and ${\alpha  = 20}$, is a scaling factor. The anomaly score is thresholded by the parameter T to obtain predicted foreground and background masks, ${\tilde m_f^q}$ and ${\tilde m_b^q}$.
% Then, a parameter T is used to threshold the anomaly score S, and the predicted foreground mask ${\tilde m_f^q}$ and background mask ${\tilde m_b^q}$ are obtained.
\begin{equation}
\tilde m_f^q = (1 - \sigma (S(x,y) - T)) \label{eq}
\end{equation}
\begin{equation}
\tilde m_b^q = 1 - \tilde m_f^q \label{eq}
\end{equation}

Where ${\sigma}$ represents a shifted sigmoid function, the subscript ${_f}$ and ${_b}$ represent the foreground and background. T is the adaptive threshold learned by the model according to the query feature ${F_q^{'}}$. T is obtained by adding a fully connected layer attached to the top of the feature extractor ${{f_\theta }}$.
% the following formula:
% \begin{equation}
% T = {h_\theta }(F_q^{'}) \label{eq}
% \end{equation}

% Where ${h_\theta}$ is a fully connected layer attached to the top of the feature extractor ${{f_\theta }}$.

\subsection{Multi-scale Prediction Fusion Module}
% The foreground and background mask predicted by the model can be divided into two paths. One path is obtained based on the 64*64 feature map, and the other is obtained based on the 32*32 feature map. In order to obtain the final prediction result, the two predicted foreground masks ${\tilde m_{64}^q}$ and ${\tilde m_{32}^q}$ are up-sampled to the size of the original figure (H,W) to obtain ${\hat m_{64}^q}$ and ${\hat m_{32}^q}$. The final prediction mask ${\hat m^q}$ is then obtained by combining with an equilibrium factor ${\alpha }$:
The model predicts foreground and background masks through two paths based on 64*64 and 32*32 feature maps. The predicted masks ${\tilde m_{64}^q}$ and ${\tilde m_{32}^q}$ are up-sampled to the original image size (H,W) to obtain ${\hat m_{64}^q}$ and ${\hat m_{32}^q}$. The final mask ${\hat m^q}$ is combined with an equilibrium factor ${\alpha}$:
% and combined with an equilibrium factor ${\alpha}$ to obtain the final mask ${\hat m^q}$:
\begin{equation}
{\hat m_q} = \alpha  \cdot {\hat m^q}_{64} + (1 - \alpha ) \cdot \hat m_{32}^q \label{eq}
\end{equation}

Where the subscripts ${_{64}}$ and ${_{32}}$ represent the dimensions are 64*64 and 32*32 respectively. ${\alpha  \in (0,1.0)}$, its value can be adjusted manually. In the experimental section, we did several experiments to get the optimal ${\alpha}$.

\subsection{Loss Function}
We used the cross entropy loss of the predicted mask and the true mask as the segmentation loss during the training stage:

\begin{align}
\begin{split}
{L_{seg}} =  - \frac{1}{{HW}}\sum\limits_{x,y} {m_f^q} (x,y)\log (\hat m_f^q(x,y)) + \\
m_b^q(x,y)\log (\hat m_b^q(x,y))\label{eq}
\end{split}
\end{align}

To align query and support images, we introduce an alignment regularization loss based on Wang et al.'s\cite{b10} proposal. It involves segmenting support images based on prototype information from query images. The alignment loss is calculated as shown below:

\begin{align}
\begin{split}
{L_{reg}} =  - \frac{1}{{HW}}\sum\limits_{x,y} {m_f^s} (x,y)\log (\hat m_f^s(x,y)) + \\
m_b^s(x,y)\log (\hat m_b^s(x,y))\label{eq}
\end{split}
\end{align}

Finally, the Loss can be expressed as:
\begin{equation}
Loss = {L_{seg}} + {L_{reg}} \label{eq}
\end{equation}

\begin{table*}[t]
\centering
\captionsetup{justification=raggedright}
\caption{Comparison of our proposed method with other methods on the CMR dataset.}
\resizebox{0.6\textwidth}{!}{
\begin{tabular}{@{}cccccc@{}}
\toprule
\textbf{Settings} & \textbf{Method} & \textbf{LV-BP}      & \textbf{LV-MYO}     & \textbf{RV}         & \textbf{Mean Dice}  \\ \midrule
Setting 1         & ALPNet\cite{b7}          & 83.99±1.43          & \textbf{66.74±3.37} & \textbf{79.96±5.89} & 76.90±3.56          \\
                  & ADNet\cite{b8}           & 86.62±1.72          & 62.41±3.35          & 74.93±4.64          & 74.65±3.24          \\
                  & Q-Net\cite{b9}           & 89.91±0.88          & 64.42±1.50          & 75.70±1.61          & 76.68±1.33          \\
                  & CRAPNet\cite{b32}         & 86.74±2.12          & 59.14±1.38          & 73.74±2.10          & 73.21±1.87          \\
                  & Proposed        & \textbf{89.95±0.83} & 66.47±1.06          & 78.15±1.46          & \textbf{78.19±1.12} \\ \bottomrule
\end{tabular}
}
\label{table1}
% \vspace{1.0em}
\end{table*}

\begin{table*}[t]
\centering
\captionsetup{justification=raggedright}
\caption{Comparison of our proposed method with other methods on the CHAOS dataset.}
\resizebox{0.6\textwidth}{!}{
\begin{tabular}{@{}ccccccc@{}}
\toprule
\textbf{Settings} & \textbf{Method} & \textbf{Liver}      & \textbf{R.kidney}   & \textbf{L.kidney}   & \textbf{Spleen}     & \textbf{Mean Dice}  \\ \midrule
Setting 1         & ALPNet\cite{b7}          & 73.98±3.03          & 83.59±2.41          & \textbf{81.92±2.93} & 71.26±7.79          & 77.69±4.04          \\
                  & ADNet\cite{b8}           & 81.88±0.77          & 79.84±14.87         & 66.24±14.92         & 71.87±13.19         & 74.96±10.94         \\
                  & Q-Net\cite{b9}           & \textbf{84.41±1.58} & 88.92±1.14          & 77.46±8.04          & 70.02±9.65          & 80.20±5.94          \\
                  & CRAPNet\cite{b32}         & 82.63±2.01          & 85.04±1.36          & 78.23±6.34          & 70.52±7.94          & 79.10±4.41          \\
                  & Proposed        & 84.05±2.12          & \textbf{89.43±0.71} & 80.72±7.79          & \textbf{74.13±6.82} & \textbf{82.08±4.36} \\
Setting 2         & ALPNet\cite{b7}          & 71.74±2.34          & \textbf{73.30±7.55} & 61.30±5.13          & 62.98±4.64          & 67.33±4.92          \\
                  & ADNet\cite{b8}           & 77.12±4.43          & 61.11±5.91          & 65.93±6.41          & 62.65±6.47          & 66.70±5.81          \\
                  & Q-Net\cite{b9}           & 78.25±4.81          & 48.98±4.11          & 55.44±4.26          & 62.57±8.73          & 61.31±5.48          \\
                  & CRAPNet\cite{b32}         & 80.09±3.52          & 57.83±3.16          & 63.64±3.48          & 60.97±7.81          & 65.63±4.49          \\
                  & Proposed        & \textbf{80.44±3.54} & 71.39±4.36          & \textbf{72.89±6.47} & \textbf{66.00±7.10} & \textbf{72.68±5.37} \\ \bottomrule
\end{tabular} }
\label{table2}
% \vspace{1.0em}
\end{table*}

\section{Results}
\subsection{Datasets}
We conducted experiments on representative public datasets CHAOS \cite{b30} and CMR \cite{b31}:

(1) The 2019 ISBI Combined Healthy Abdominal Organ Segmentation Challenge (CHAOS) task 5 was launched for abdominal organ segmentation. 20 3D T2 sequence scans from the liver, left kidney, right kidney, and spleen were included, with each scan averaging 36 slices.

(2) The MICCAAI 2019 Multi-sequence Heart MRI Segmentation Challenge published the CMR dataset, which was an MRI dataset of heart organ segmentation. It consisted of 35 heart MRI scans, each with around 13 slices.

To ensure fairness, we used the same pre-processing scheme as Ouyang \cite{b7}, which involved deleting the bright end of the histogram to solve resonance issues, resampling images to the same resolution, and unifying all images to 256x256 pixels. Additionally, since the pre-training network required three-channel input, we repeated each single-channel medical image slice three times in the channel dimension.

Models undergo five-fold cross-validation after self-supervised training using supervoxel-based pseudo labels created offline. Minimum supervoxel size is controlled by parameter during supervoxel generation, affecting segmentation outcome. CHAOS set to 5000 and CMR set to 1000 based on Hansen's advice. Three-dimensional image is sliced and segmented during evaluation with one volume chosen as support and the rest as queries. Each fold is repeated three times to account for randomness.

\subsection{Evaluation Metric}
We used the mean dice score as the evaluation criterion in line with standard medical segmentation procedures. It contrasts the projected segmentation A with the actual segmentation B's overlap ratio. The equation reads as follows:
\begin{equation}
{\rm{D}}(A,B) = 2\frac{{|A \cap B|}}{{|A| + |B|}} \cdot 100\% \label{eq}
\end{equation}
Where the higher the Dice score, the better the segmentation effect.

\begin{figure*}[t]
\setlength{\belowcaptionskip}{5mm}
\centering
\subfigure{
\begin{minipage}[t]{0.18\textwidth}
\centering
{\scriptsize{ADNet\cite{b8}}}
\end{minipage}
\begin{minipage}[t]{0.18\textwidth}
\centering
{\scriptsize{Q-Net\cite{b9}}}
\end{minipage}
\begin{minipage}[t]{0.18\textwidth}
\centering
{\scriptsize{CRAPNet\cite{b32}}}
\end{minipage}
\begin{minipage}[t]{0.18\textwidth}
\centering
{\scriptsize{Proposed}}
\end{minipage}
\begin{minipage}[t]{0.18\textwidth}
\centering
{\scriptsize{GT}}
\end{minipage}
}

% {\rotatebox{90}{\scriptsize{~~~~~~~~LV-MYO }}}
\subfigure{
\begin{minipage}[t]{0.18\textwidth}
\centering
\includegraphics[width=\textwidth]{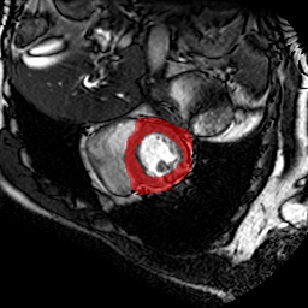}
\end{minipage}%
}%
\subfigure{
\begin{minipage}[t]{0.18\textwidth}
\centering
\includegraphics[width=\textwidth]{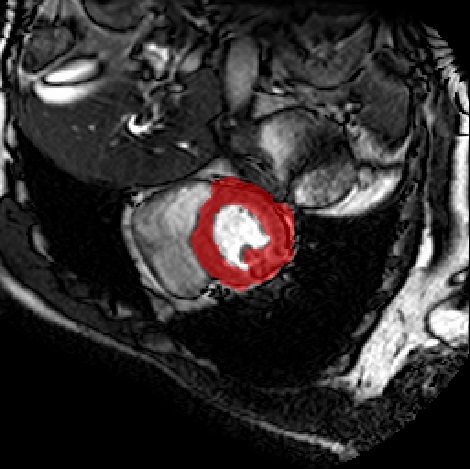}
\end{minipage}%
}%
\subfigure{
\begin{minipage}[t]{0.18\textwidth}
\centering
\includegraphics[width=\textwidth]{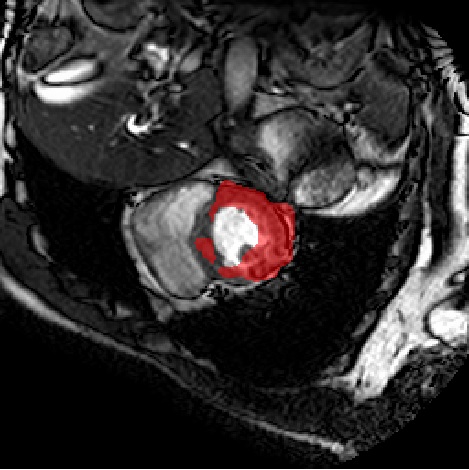}
\end{minipage}%
}%
\subfigure{
\begin{minipage}[t]{0.18\textwidth}
\centering
\includegraphics[width=\textwidth]{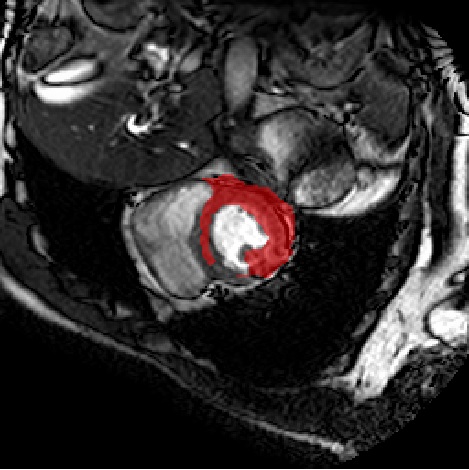}
\end{minipage}
}%
\subfigure{
\begin{minipage}[t]{0.18\textwidth}
\centering
\includegraphics[width=\textwidth]{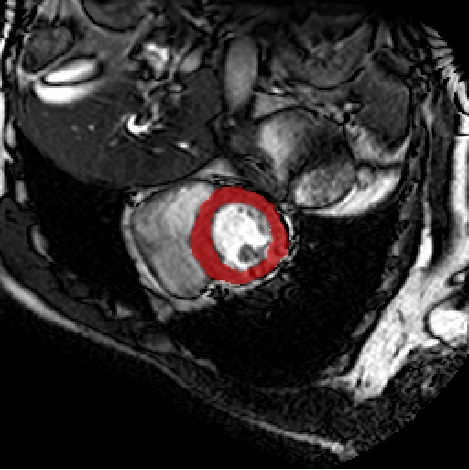}
\end{minipage}%
}%

% {\rotatebox{90}{\scriptsize{~~~~~~~~~LV-BP }}}
\subfigure{
\begin{minipage}[t]{0.18\textwidth}
\centering
\includegraphics[width=\textwidth]{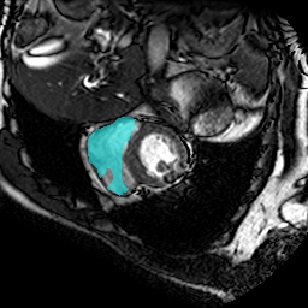}
\end{minipage}%
}%
\subfigure{
\begin{minipage}[t]{0.18\textwidth}
\centering
\includegraphics[width=\textwidth]{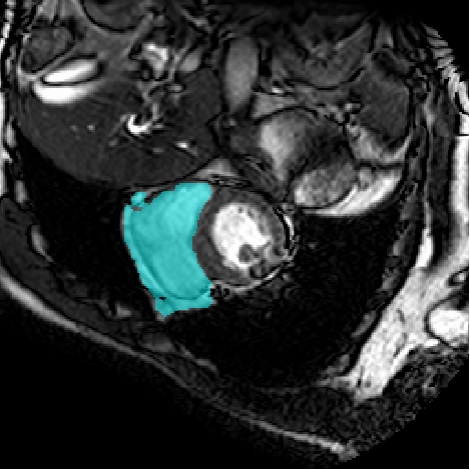}
\end{minipage}%
}%
\subfigure{
\begin{minipage}[t]{0.18\textwidth}
\centering
\includegraphics[width=\textwidth]{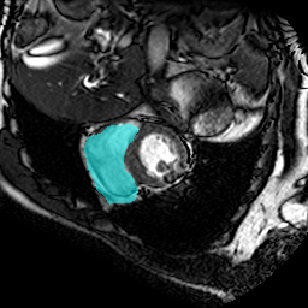}
\end{minipage}%
}%
\subfigure{
\begin{minipage}[t]{0.18\textwidth}
\centering
\includegraphics[width=\textwidth]{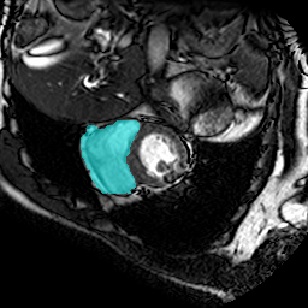}
\end{minipage}%
}%
\subfigure{
\begin{minipage}[t]{0.18\textwidth}
\centering
\includegraphics[width=\textwidth]{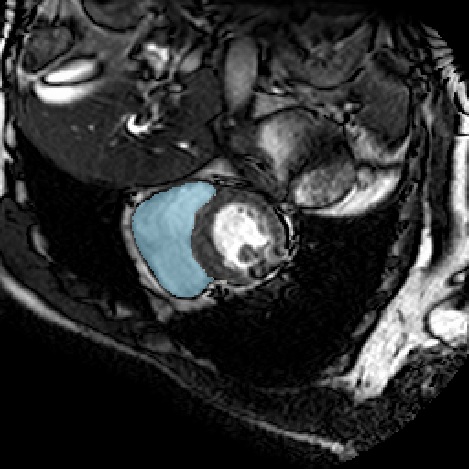}
\end{minipage}%
}%

% {\rotatebox{90}{\scriptsize{~~~~~~~~~~~~RV }}}
\subfigure{
\begin{minipage}[t]{0.18\textwidth}
\centering
\includegraphics[width=\textwidth]{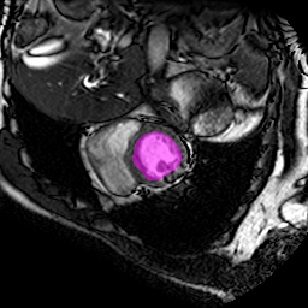}
\end{minipage}%
}%
\subfigure{
\begin{minipage}[t]{0.18\textwidth}
\centering
\includegraphics[width=\textwidth]{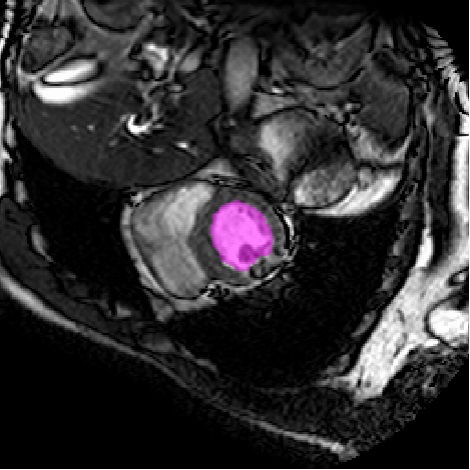}
\end{minipage}%
}%
\subfigure{
\begin{minipage}[t]{0.18\textwidth}
\centering
\includegraphics[width=\textwidth]{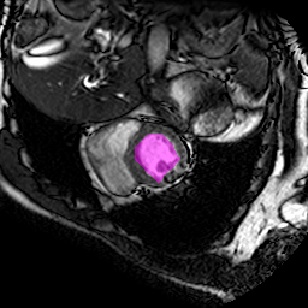}
\end{minipage}%
}%
\subfigure{
\begin{minipage}[t]{0.18\textwidth}
\centering
\includegraphics[width=\textwidth]{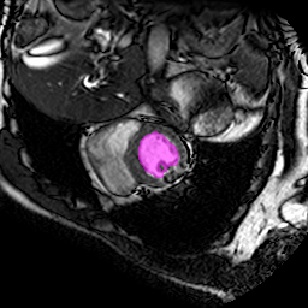}
\end{minipage}%
}%
\subfigure{
\begin{minipage}[t]{0.18\textwidth}
\centering
\includegraphics[width=\textwidth]{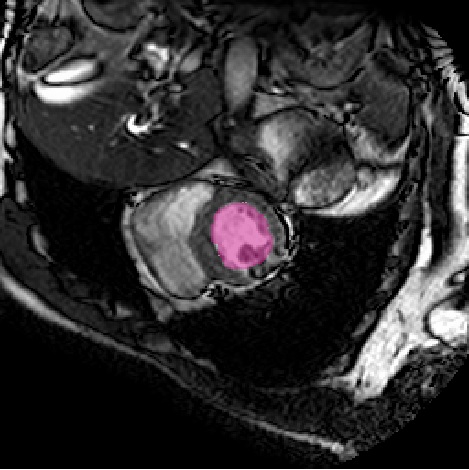}
\end{minipage}%
}%
\captionsetup{justification=raggedright}
\caption{Comparison of segmentation results from Setting1 on CMR dataset. From left to right: ADNet, Q-Net, CRAPNet, our proposed method and GT. From top to bottom: LV-MYO, LY-BP and RV.}
\label{fig4}
\end{figure*}

\begin{figure*}[!t]
\setlength{\belowcaptionskip}{5mm}
\centering
\subfigure{
\begin{minipage}[t]{0.16\textwidth}
\centering
{\scriptsize{ALPNet\cite{b7}}}
\end{minipage}
\begin{minipage}[t]{0.16\textwidth}
\centering
{\scriptsize{ADNet\cite{b8}}}
\end{minipage}
\begin{minipage}[t]{0.16\textwidth}
\centering
{\scriptsize{Q-Net\cite{b9}}}
\end{minipage}
\begin{minipage}[t]{0.16\textwidth}
\centering
{\scriptsize{CRAPNet\cite{b32}}}
\end{minipage}
\begin{minipage}[t]{0.16\textwidth}
\centering
{\scriptsize{Proposed}}
\end{minipage}
\begin{minipage}[t]{0.16\textwidth}
\centering
{\scriptsize{GT}}
\end{minipage}
}

% {\rotatebox{90}{\scriptsize{~~~~~~~~~~~~~~~Liver}}}
\subfigure{
\begin{minipage}[t]{0.16\textwidth}
\centering
\includegraphics[width=1in]{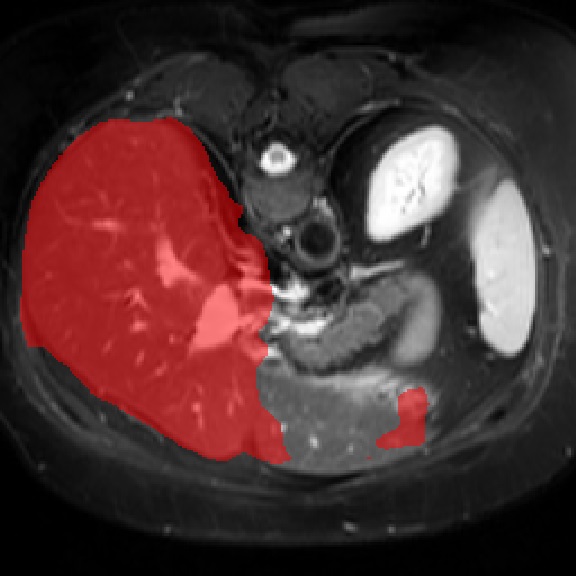}
%\caption{fig1}
\end{minipage}%
}%
\subfigure{
\begin{minipage}[t]{0.16\textwidth}
\centering
\includegraphics[width=1in]{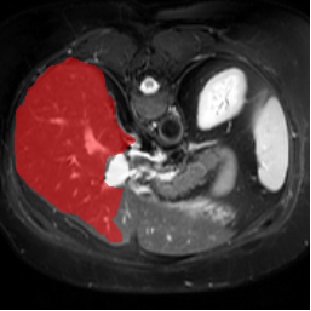}
%\caption{fig1}
\end{minipage}%
}%
\subfigure{
\begin{minipage}[t]{0.16\textwidth}
\centering
\includegraphics[width=1in]{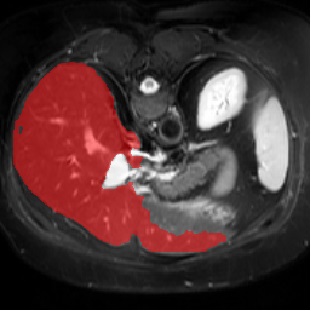}
%\caption{fig1}
\end{minipage}%
}%
\subfigure{
\begin{minipage}[t]{0.16\textwidth}
\centering
\includegraphics[width=1in]{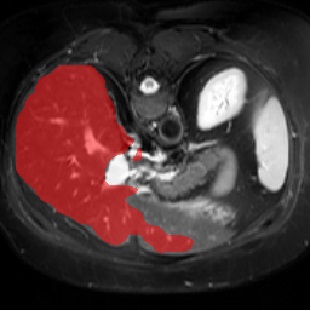}
%\caption{fig1}
\end{minipage}%
}%
\subfigure{
\begin{minipage}[t]{0.16\textwidth}
\centering
\includegraphics[width=1in]{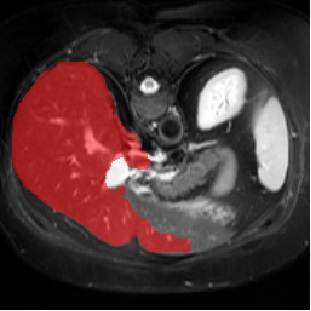}
%\caption{fig1}
\end{minipage}%
}%
\subfigure{
\begin{minipage}[t]{0.16\textwidth}
\centering
\includegraphics[width=1in]{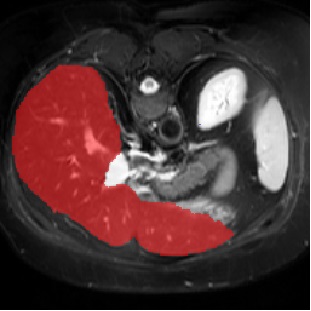}
%\caption{fig1}
\end{minipage}%
}%

% {\rotatebox{90}{\scriptsize{~~~~~~~~~~~R.kidney}}}
\subfigure{
\begin{minipage}[t]{0.16\textwidth}
\centering
\includegraphics[width=1in]{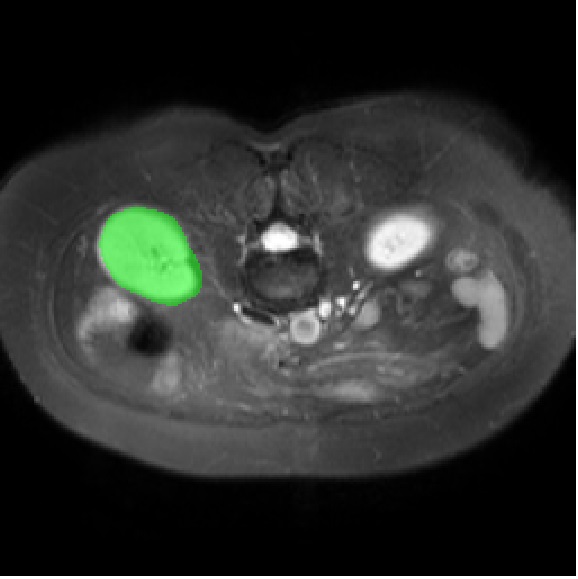}
%\caption{fig1}
\end{minipage}%
}%
\subfigure{
\begin{minipage}[t]{0.16\textwidth}
\centering
\includegraphics[width=1in]{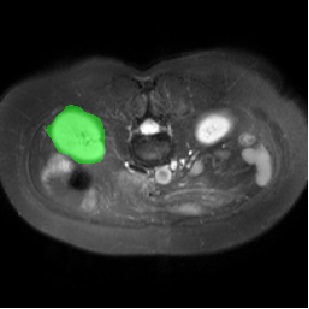}
%\caption{fig1}
\end{minipage}%
}%
\subfigure{
\begin{minipage}[t]{0.16\textwidth}
\centering
\includegraphics[width=1in]{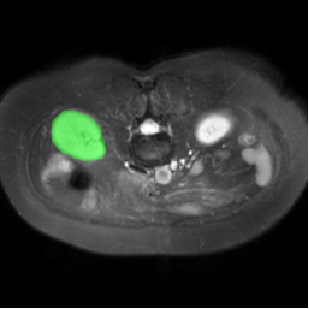}
%\caption{fig1}
\end{minipage}%
}%
\subfigure{
\begin{minipage}[t]{0.16\textwidth}
\centering
\includegraphics[width=1in]{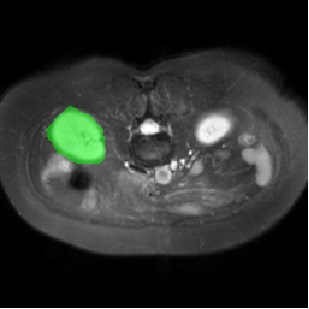}
%\caption{fig1}
\end{minipage}%
}%
\subfigure{
\begin{minipage}[t]{0.16\textwidth}
\centering
\includegraphics[width=1in]{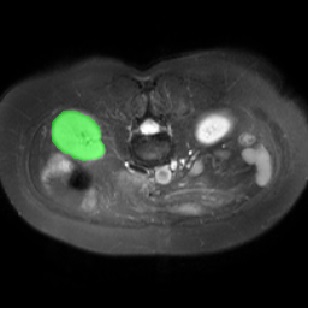}
%\caption{fig1}
\end{minipage}%
}%
\subfigure{
\begin{minipage}[t]{0.16\textwidth}
\centering
\includegraphics[width=1in]{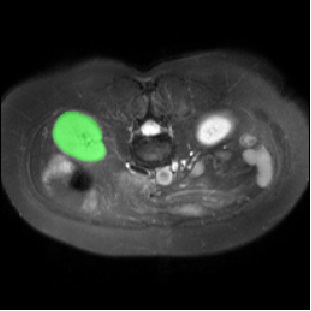}
%\caption{fig1}
\end{minipage}%
}%

% {\rotatebox{90}{\scriptsize{~~~~~~~~~~~L.kidney}}}
\subfigure{
\begin{minipage}[t]{0.16\textwidth}
\centering
\includegraphics[width=1in]{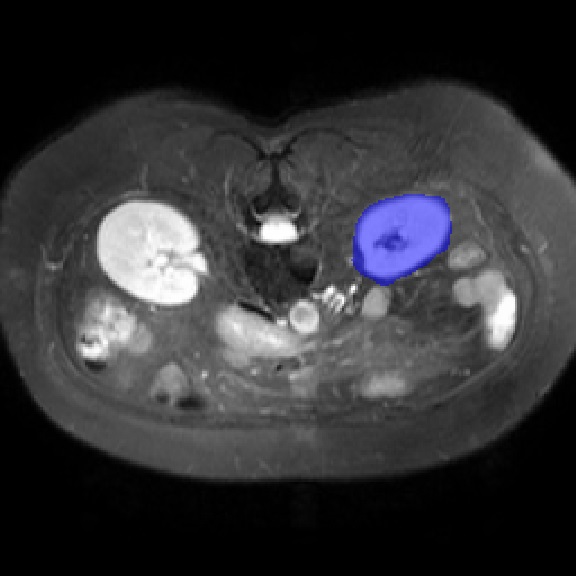}
%\caption{fig1}
\end{minipage}%
}%
\subfigure{
\begin{minipage}[t]{0.16\textwidth}
\centering
\includegraphics[width=1in]{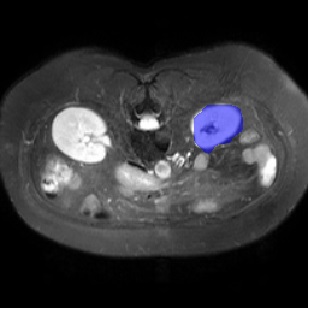}
%\caption{fig1}
\end{minipage}%
}%
\subfigure{
\begin{minipage}[t]{0.16\textwidth}
\centering
\includegraphics[width=1in]{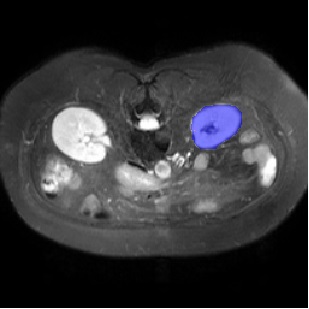}
%\caption{fig1}
\end{minipage}%
}%
\subfigure{
\begin{minipage}[t]{0.16\textwidth}
\centering
\includegraphics[width=1in]{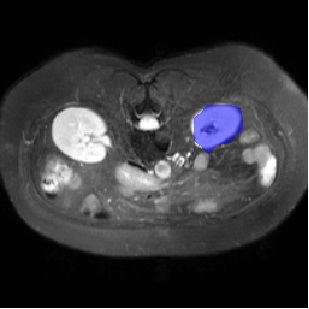}
%\caption{fig1}
\end{minipage}%
}%
\subfigure{
\begin{minipage}[t]{0.16\textwidth}
\centering
\includegraphics[width=1in]{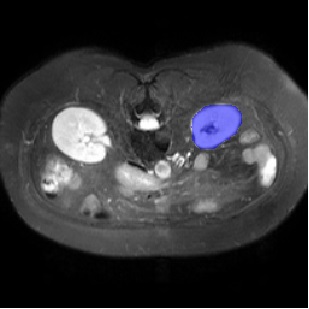}
%\caption{fig1}
\end{minipage}%
}%
\subfigure{
\begin{minipage}[t]{0.16\textwidth}
\centering
\includegraphics[width=1in]{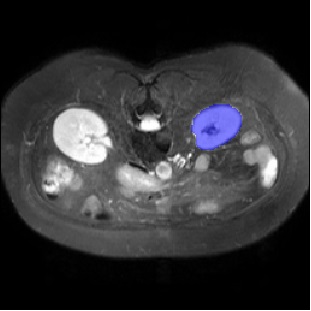}
%\caption{fig1}
\end{minipage}%
}%

% {\rotatebox{90}{\scriptsize{~~~~~~~~~~~~~~Spleen}}}
\subfigure{
\begin{minipage}[t]{0.16\textwidth}
\centering
\includegraphics[width=1in]{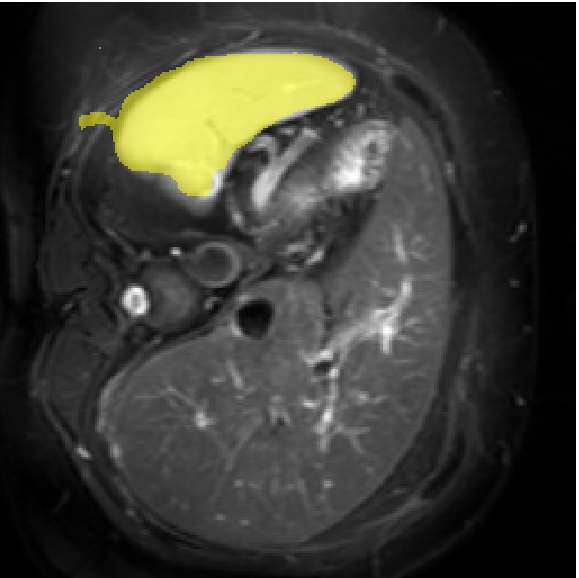}
%\caption{fig1}
\end{minipage}%
}%
\subfigure{
\begin{minipage}[t]{0.16\textwidth}
\centering
\includegraphics[width=1in]{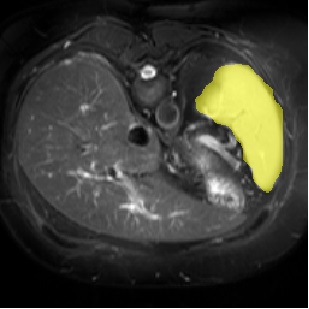}
%\caption{fig1}
\end{minipage}%
}%
\subfigure{
\begin{minipage}[t]{0.16\textwidth}
\centering
\includegraphics[width=1in]{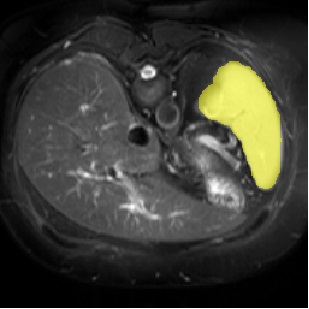}
%\caption{fig1}
\end{minipage}%
}%
\subfigure{
\begin{minipage}[t]{0.16\textwidth}
\centering
\includegraphics[width=1in]{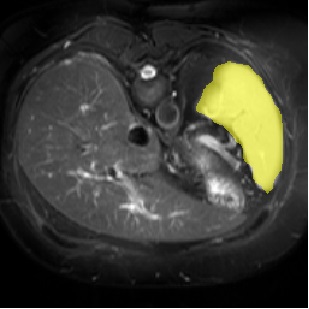}
%\caption{fig1}
\end{minipage}%
}%
\subfigure{
\begin{minipage}[t]{0.16\textwidth}
\centering
\includegraphics[width=1in]{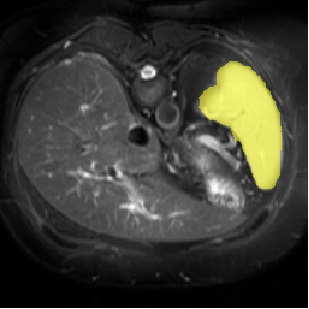}
%\caption{fig1}
\end{minipage}%
}%
\subfigure{
\begin{minipage}[t]{0.16\textwidth}
\centering
\includegraphics[width=1in]{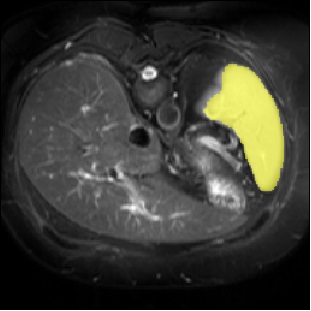}
%\caption{fig1}
\end{minipage}%
}%
\captionsetup{justification=raggedright}
\caption{Comparison of segmentation results from Setting1 on CHAOS dataset. From left to right: ALPNet, ADNet, Q-Net, CRAPNet, proposed model and GT. From top to bottom: Liver, R.kidney, L.kidney, and Spleen. }
\label{fig5}
\end{figure*}

\begin{figure*}[!t]
\setlength{\belowcaptionskip}{5mm}
\centering
\subfigure{
\begin{minipage}[t]{0.16\textwidth}
\centering
{\scriptsize{ALPNet\cite{b7}}}
\end{minipage}
\begin{minipage}[t]{0.16\textwidth}
\centering
{\scriptsize{ADNet\cite{b8}}}
\end{minipage}
\begin{minipage}[t]{0.16\textwidth}
\centering
{\scriptsize{Q-Net\cite{b9}}}
\end{minipage}
\begin{minipage}[t]{0.16\textwidth}
\centering
{\scriptsize{CRAPNet\cite{b32}}}
\end{minipage}
\begin{minipage}[t]{0.16\textwidth}
\centering
{\scriptsize{Proposed}}
\end{minipage}
\begin{minipage}[t]{0.16\textwidth}
\centering
{\scriptsize{GT}}
\end{minipage}
}

% {\rotatebox{90}{\scriptsize{~~~~~~~~~~~~~~Liver}}}
\subfigure{
\begin{minipage}[t]{0.16\textwidth}
\centering
\includegraphics[width=1in]{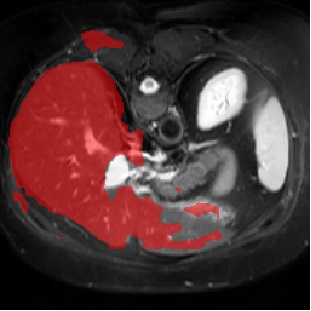}
%\caption{fig1}
\end{minipage}%
}%
\subfigure{
\begin{minipage}[t]{0.16\textwidth}
\centering
\includegraphics[width=1in]{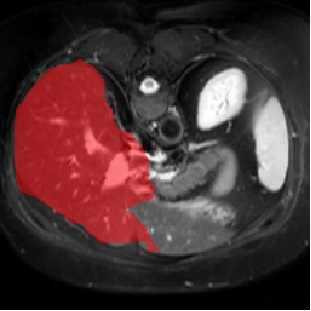}
%\caption{fig1}
\end{minipage}%
}%
\subfigure{
\begin{minipage}[t]{0.16\textwidth}
\centering
\includegraphics[width=1in]{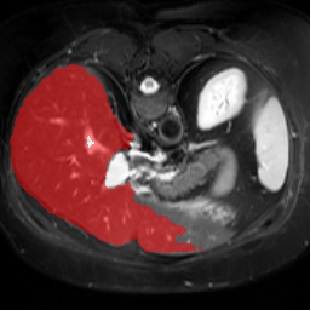}
%\caption{fig1}
\end{minipage}%
}%
\subfigure{
\begin{minipage}[t]{0.16\textwidth}
\centering
\includegraphics[width=1in]{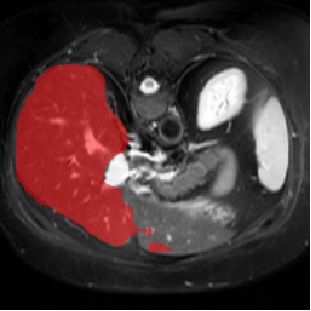}
%\caption{fig1}
\end{minipage}%
}%
\subfigure{
\begin{minipage}[t]{0.16\textwidth}
\centering
\includegraphics[width=1in]{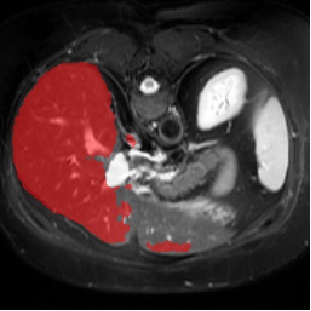}
%\caption{fig1}
\end{minipage}%
}%
\subfigure{
\begin{minipage}[t]{0.16\textwidth}
\centering
\includegraphics[width=1in]{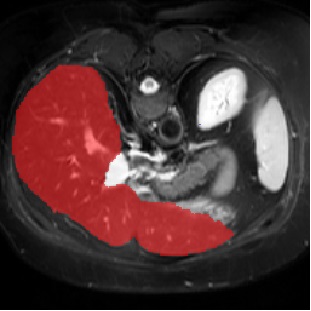}
%\caption{fig1}
\end{minipage}%
}%

% {\rotatebox{90}{\scriptsize{~~~~~~~~~~~~~~R.kidney}}}
\subfigure{
\begin{minipage}[t]{0.16\textwidth}
\centering
\includegraphics[width=1in]{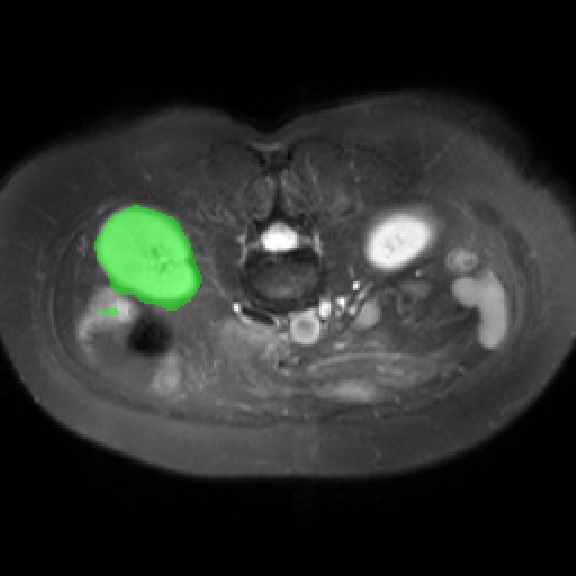}
%\caption{fig1}
\end{minipage}%
}%
\subfigure{
\begin{minipage}[t]{0.16\textwidth}
\centering
\includegraphics[width=1in]{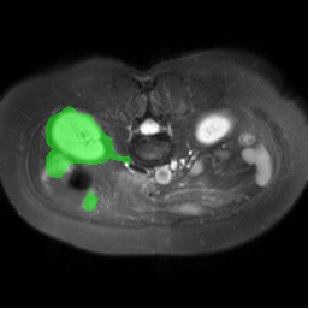}
%\caption{fig1}
\end{minipage}%
}%
\subfigure{
\begin{minipage}[t]{0.16\textwidth}
\centering
\includegraphics[width=1in]{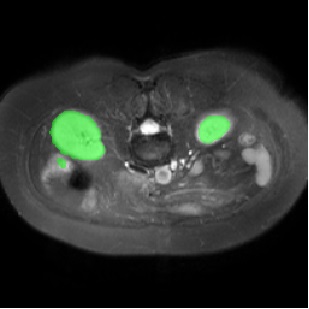}
%\caption{fig1}
\end{minipage}%
}%
\subfigure{
\begin{minipage}[t]{0.16\textwidth}
\centering
\includegraphics[width=1in]{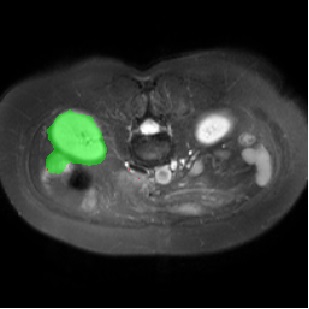}
%\caption{fig1}
\end{minipage}%
}%
\subfigure{
\begin{minipage}[t]{0.16\textwidth}
\centering
\includegraphics[width=1in]{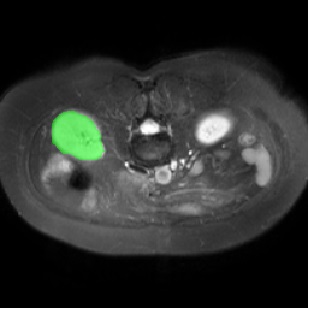}
%\caption{fig1}
\end{minipage}%
}%
\subfigure{
\begin{minipage}[t]{0.16\textwidth}
\centering
\includegraphics[width=1in]{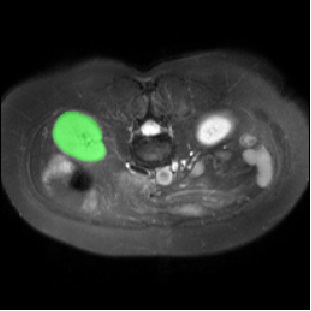}
%\caption{fig1}
\end{minipage}%
}%

% {\rotatebox{90}{\scriptsize{~~~~~~~~~~~~~~L.kidney}}}
\subfigure{
\begin{minipage}[t]{0.16\textwidth}
\centering
\includegraphics[width=1in]{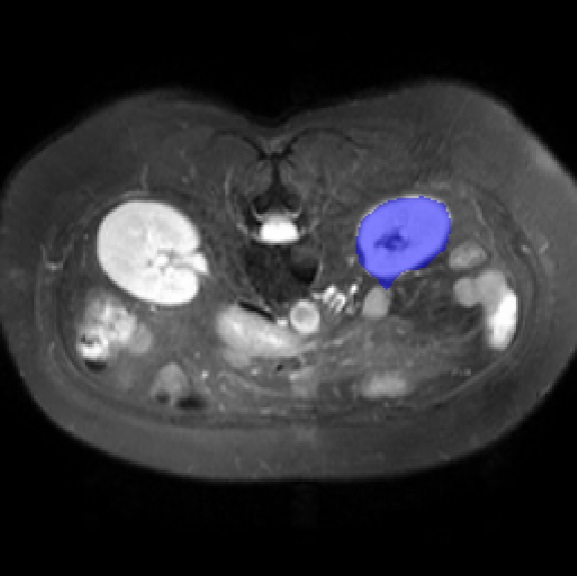}
%\caption{fig1}
\end{minipage}%
}%
\subfigure{
\begin{minipage}[t]{0.16\textwidth}
\centering
\includegraphics[width=1in]{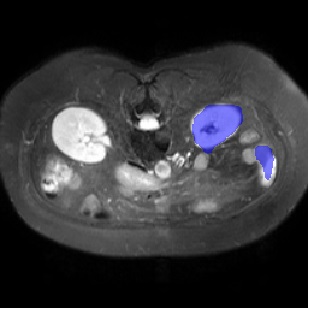}
%\caption{fig1}
\end{minipage}%
}%
\subfigure{
\begin{minipage}[t]{0.16\textwidth}
\centering
\includegraphics[width=1in]{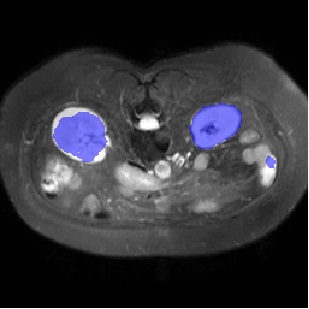}
%\caption{fig1}
\end{minipage}%
}%
\subfigure{
\begin{minipage}[t]{0.16\textwidth}
\centering
\includegraphics[width=1in]{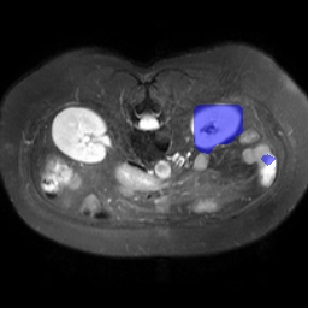}
%\caption{fig1}
\end{minipage}%
}%
\subfigure{
\begin{minipage}[t]{0.16\textwidth}
\centering
\includegraphics[width=1in]{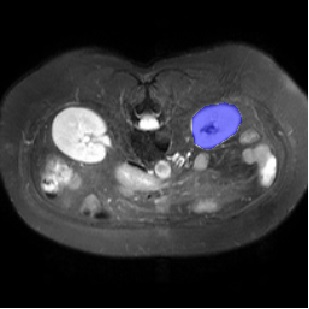}
%\caption{fig1}
\end{minipage}%
}%
\subfigure{
\begin{minipage}[t]{0.16\textwidth}
\centering
\includegraphics[width=1in]{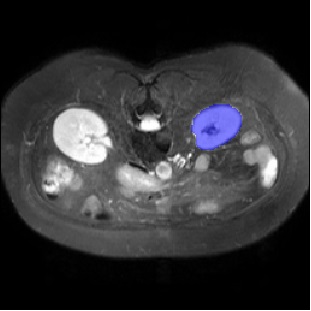}
%\caption{fig1}
\end{minipage}%
}%

% {\rotatebox{90}{\scriptsize{~~~~~~~~~~~~~~Spleen}}}
\subfigure{
\begin{minipage}[t]{0.16\textwidth}
\centering
\includegraphics[width=1in]{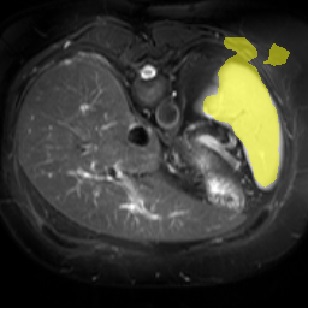}
%\caption{fig1}
\end{minipage}%
}%
\subfigure{
\begin{minipage}[t]{0.16\textwidth}
\centering
\includegraphics[width=1in]{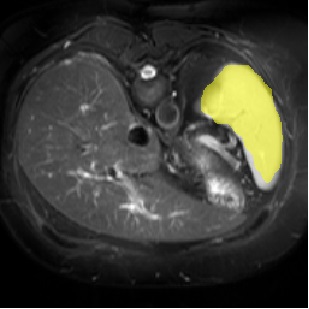}
%\caption{fig1}
\end{minipage}%
}%
\subfigure{
\begin{minipage}[t]{0.16\textwidth}
\centering
\includegraphics[width=1in]{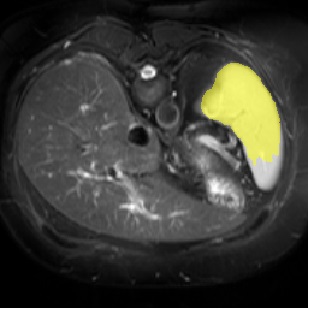}
%\caption{fig1}
\end{minipage}%
}%
\subfigure{
\begin{minipage}[t]{0.16\textwidth}
\centering
\includegraphics[width=1in]{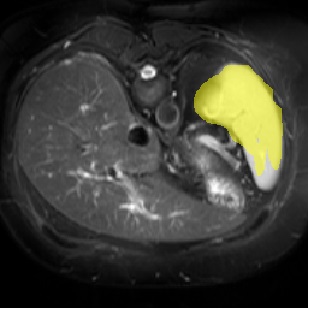}
%\caption{fig1}
\end{minipage}%
}%
\subfigure{
\begin{minipage}[t]{0.16\textwidth}
\centering
\includegraphics[width=1in]{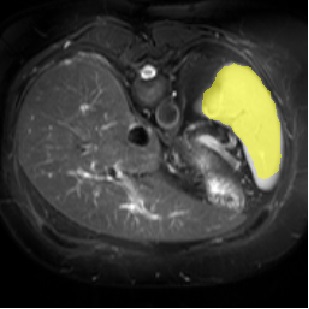}
%\caption{fig1}
\end{minipage}%
}%
\subfigure{
\begin{minipage}[t]{0.16\textwidth}
\centering
\includegraphics[width=1in]{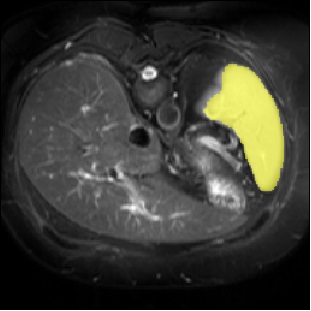}
%\caption{fig1}
\end{minipage}%
}%
\captionsetup{justification=raggedright}
\caption{Comparison of segmentation results from Setting2 on CHAOS dataset. From left to right: ALPNet, ADNet, Q-Net, CRAPNet, proposed model and GT. From top to bottom: Liver, R.kidney, L.kidney, and Spleen.}
\label{fig6}
    
\end{figure*}

\subsection{Comparison to state-of-the-art}
We compare our method with the following advanced models: ALPNet\cite{b7}, ADNet\cite{b8}, Q-Net\cite{b9} and CRAPNet\cite{b32}. Table \ref{table1} and Table \ref{table2} display the results.

To evaluate the performance of the model, two settings were used. In setting 1, test class objects may already appear in the training set's background as supervoxels unavoidably contain various information. Thus, the test classes in setting 1 are not entirely “unseen” and evaluation was higher than in setting 2. In setting 2, to ensure the test classes are truly unseen, the slice containing their images was deleted from the training set. The abdominal organ segmentation dataset was split into upper and lower abdomen organs, following Ouyang's approach.

For CMR dataset, we only consider setting 1 because setting 2 is difficult to implement. The results are shown in Table \ref{table1} and  Figure \ref{fig4}. In general, our proposed method is superior than others. For CHAOS dataset, we consider both setting 1 and setting 2. From the results, we can see the performance of our proposed model is higher than others on CHAOS and CMR datasets. Compared with Q-Net, our mean dice score on CHAOS dataset achieves about 82\%. The segmentation results of the model are shown in Figure \ref{fig5}. In setting 2, the slices of the objects containing the test classes are all removed at training stage, and our model still performs well. As can be seen from the Table \ref{table2}, the segmentation effect of the model on the right kidney is slightly worse than that of ALPNet, but overall, we achieve the best performance. The specific segmentation results are shown in Figure \ref{fig6}. As can be seen from the figure, other methods have excessive segmentation, but ours does not. Our segmentation results are more accurate and reduce redundant segmentation.

\begin{table*}[t]
\centering
\captionsetup{justification=raggedright}
\caption{Ablation study of different modules on CHAOS dataset under setting 2. DP is  Dual-path extractor, and AM is an Attention Module.}
\resizebox{0.6\linewidth}{!}{
\begin{tabular}{@{}ccccccc@{}}
\toprule
\textbf{Experiment} & \textbf{Module} & \textbf{Liver} & \textbf{R.kidney} & \textbf{L.kidney} & \textbf{Spleen} & \textbf{Mean Dice} \\ \midrule
1                   & 32×32           & 78.67          & 67.62             & 64.38             & 62.20           & 68.22±5.19         \\
2                   & 64×64           & 79.07          & 68.32             & 65.69             & 63.35           & 69.11±8.36         \\
3                   & DP              & 79.84          & 69.73             & 69.15             & 64.74           & 70.87±7.65         \\
4                   & DP+AM           & 80.44          & 71.39             & 72.89             & 66.00           & 72.68±6.63         \\ \bottomrule
\end{tabular}
}
\label{table3}
% \vspace{1.0em}
\end{table*}

\begin{table}[t]
\centering
\captionsetup{justification=raggedright}
\caption{Ablation study of ${\alpha}$.}
\begin{tabular}{@{}cc@{}}
\toprule
\textbf{${\alpha}$} & \textbf{Setting2} \\ \midrule
0.9           & 72.21±7.06        \\
0.8           & 72.68±6.63        \\
0.6           & 67.36±6.82        \\
0.4           & 72.30±7.40        \\
0.2           & 66.70±8.68        \\ \bottomrule
\end{tabular}
\label{table4}
% \vspace{1.0em}
\end{table}

\subsection{Ablation Study}
Firstly, we conduct experiments to verify the effectiveness of the dual-path feature extractor, the attention module and the adaptive prototype prediction module. We use CHAOS dataset to conduct the experiment in setting 2. The results are shown in Table \ref{table3}. As you can see from the table, each module does contribute to improving the final performance of the model.

Then, we test the effect of the balance factor ${\alpha}$ on the model. At the last layer of the model, namely the multi-scale prediction fusion layer, the prediction results obtained from the feature maps of different sizes will be fused. Where, ${\alpha}$ represents the proportion of mask predicted according to the feature map size 64*64 in the final prediction result. Considering the different sizes of the partitioned organs, which may lead to different results of the model, we conducted many experiments to verify the effect under setting 2, and the results are shown in Table \ref{table4}. As we can see from the table, the optimal value is 0.8. Specifically, the model works well when ${\alpha}$ is 0.4 or 0.8, and 0.8 has the best performance. Of the four organs to be segmented, the upper abdominal organs (liver and spleen) are both larger, while the lower abdominal organs (left and right kidneys) are smaller. As the value of ${\alpha}$ rises, the segmentation impact of the upper abdominal organs is enhanced by the model's usage of a larger segmentation mask, while the segmentation effect of the lower abdominal organs is diminished. The model works best at balancing different organs when ${\alpha}$ is 0.8. Besides, the model's performance improved by setting the kernel size of the attention module's DW convolution and DW-D convolution to 3, instead of the original 5 based on guo et al.'s\cite{b29} experience. Large scale convolution caused the model to focus excessively on larger organs like the liver, leading to a good performance in the upper abdomen but a poor performance in the lower abdomen, ultimately affecting the overall performance negatively.

\section{Conclusion}
We presented a plug-and-play attention module utilizing a prototype network technique based on the work of Shen et al. This attention module is capable of adaptive enhancement of support features and query features to capture not only local features but also long-range features to obtain comprehensive feature representation. In general, our model starts from the aspects of multi-scale and multi-dimension, making the extracted prototype more representative and achieving the most advanced performance on the CHAOS and CMR datasets. However, some difficulties still exist. First, attention mechanisms may not be as effective for fine segmentation since they frequently concentrate on the more crucial components. For smaller organs and the division of organ margins, for instance, using attention mechanisms alone is insufficient. Secondly, for the few-shot image segmentation, the most urgent problem to be solved should be the scarcity of medical annotated data. Subsequent work could consider data enhancement strategies or the use of GANs to generate high quality images. In addition, Data Heterogeneity, Model Interpretability and Generalizability and Robustness are also important issues. Addressing these challenges requires collaboration between researchers, hospitals, and other institutions.

\section{Declarations}
\subsection{Conflict of interest}
The authors declare that they have no known competing financial interests or personal relationships that could have appeared to influence the work reported in this paper.
\subsection{Ethical Approval}
Not applicable.
\subsection{Funding}
No applicable.
\subsection{Availability of data and materials}
The datasets we use are all public open-source data,
they can be obtained from \cite{b30,b31}

% following only if there is an appendix
% \section*{Appendix}
% \addcontentsline{toc}{section}{\numberline{}Appendix}
% Appendix text goes here if needed.

% \section*{References}
% \addcontentsline{toc}{section}{\numberline{}References}
% \vspace*{-20mm}


\begin{thebibliography}{00}


\bibitem{b1} Chen L C, Papandreou G, Kokkinos I, et al. Deeplab: Semantic image segmentation with deep convolutional nets, atrous convolution, and fully connected crfs. \emph{IEEE transactions on pattern analysis and machine intelligence}. 2018; 40(4): 834–848. \href{https://arxiv.org/abs/1606.00915v2}{https://arxiv.org/abs/1606.00915v2}

\bibitem{b2} Zhao H, Shi J, Qi X, et al. Pyramid scene parsing network. in \emph{Proceedings of the IEEE conference on computer vision and pattern recognition.} 2017, pp. 2881-2890. \href{https://ieeexplore.ieee.org/document/8100143}{https://ieeexplore.ieee.org/document/8100143}

\bibitem{b3} Long J, Shelhamer E, and Darrell T, Fully convolutional networks for semantic segmentation. in \emph{Proceedings of the IEEE conference on computer vision and pattern recognition,} 2015, pp. 3431–3440. \href{https://arxiv.org/abs/1411.4038}{https://arxiv.org/abs/1411.4038}

\bibitem{b4} Lin G, Milan A, Shen C, et al. Refinenet: Multi-path refinement networks for high-resolution semantic segmentation. in \emph{Proceedings of the IEEE conference on computer vision and pattern recognition.} 2017, pp. 1925-1934. \href{https://arxiv.org/abs/1611.06612v3}{https://arxiv.org/abs/1611.06612v3}

\bibitem{b5} Byoung Chul Ko, JiHyeon Lee, and Jae-Yeal Nam. Automatic medical image annotation and keyword-based image retrieval using relevance feedback. \emph{Journal of digital imaging}, 2012; 25(4): 454–465. \href{https://link.springer.com/article/10.1007/s10278-011-9443-5}{https://link.springer.com/article/10.1007/s102\\78-011-9443-5}

\bibitem{b6} Samuel Budd, Emma C Robinson, and Bernhard Kainz. A survey on active learning and human-in-the-loop deep learning for medical image analysis. \emph{Medical Image Analysis}, 2021; 71: 102062. \href{https://arxiv.org/abs/1910.02923}{https://arxiv.org/abs/1910.02923}

\bibitem{b7} Ouyang C, Biffi C, Chen C, et al, Self-supervision with superpixels: Training few-shot medical image segmentation without annotation. in \emph{European Conference on Computer Vision.} Springer, 2020, pp. 762-780. \href{https://arxiv.org/abs/2007.09886}{https://arxiv.org/abs/2007.09886}


\bibitem{b8} Hansen S, Gautam S, Jenssen R, et al, Anomaly detection-inspired few-shot medical image segmentation through self-supervision with supervoxels. \emph{Medical Image Analysis,} 2022; 78:102385. \href{https://arxiv.org/abs/2203.02048v1}{https://arxiv.org/abs/2203.02048v1}


\bibitem{b9} Shen Q, Li Y, Jin J, et al. Q-Net: Query-Informed Few-Shot Medical Image Segmentation. \emph{arXiv preprint arXiv:2208.11451}, 2022. \href{https://arxiv.org/abs/2208.11451}{https://arxiv.org/abs/2208.11451}


\bibitem{b10} Wang K, Liew J. H, Zou Y, et al, PANet: Few-shot image semantic segmentation with prototype alignment, in \emph{IEEE International Conference on Computer Vision,} 2019, pp. 9197–9206. \href{https://arxiv.org/abs/1908.06391}{https://arxiv.org/abs/1908.06391}


\bibitem{b11} Protonotarios N E, Katsamenis I, Sykiotis S, et al. A few-shot U-Net deep learning model for lung cancer lesion segmentation via PET/CT imaging. \emph{Biomedical Physics \& Engineering Express}, 2022; 8(2): 025019. \href{https://iopscience.iop.org/article/10.1088/2057-1976/ac53bd}{https://iopscience.iop.org/article/10.1088/2057-1976/ac53bd}

\bibitem{b12} Li C, Zhang D, Tian Z. Q, et al. Few-shot learning with deformable convolution for multiscale lesion detection in mammography. \emph{Medical physics}, 2020; 47: 7. \href{https://pubmed.ncbi.nlm.nih.gov/32160321/}{https://pubmed.ncbi.nlm.nih.gov/32160321/}

\bibitem{b13} Huang Q, Huang Y, Luo Y, et al, Segmentation of breast ultrasound image with semantic classification of superpixels, \emph{Medical image analysis,} 2020; 61: 101657. \href{https://pubmed.ncbi.nlm.nih.gov/32032899/}{https://pubmed.ncbi.nlm.nih.gov/32032899/}

\bibitem{b14} Irving B, Franklin J. M, Papież B. W, et al, Pieces-of-parts for supervoxel segmentation with global context: Application to DCE-MRI tumour delineation, \emph{Medical image analysis,} 2016; 32: 69-83. \href{https://arxiv.org/pdf/1604.05210v1.pdf}{https://arxiv.org/pdf/1604.05210v1.pdf}

\bibitem{b15} Stutz D, Hermans A, and Leibe B, Superpixels: An evaluation of the state-of-the-art, \emph{Computer Vision and Image Understanding,} 2018; 166: 1–27. \href{https://arxiv.org/abs/1612.01601}{https://arxiv.org/abs/1612.01601}

\bibitem{b16} Yang X. H, Wang B. R, Zhou X. C, et al, BriNet: Towards Bridging the Intra-class and Inter-class Gaps in One-Shot Segmentation, \emph{31st British Machine Vision Conference,} 2020. \href{https://doi.org/10.48550/arXiv.2008.06226}{https://doi.org/10.48550/arXiv.2008.06226}

\bibitem{b17} Woo S, Park J, Lee J. Y, et al, CBAM: Convolutional block attention module, in \emph{Proceedings of the European conference on computer vision ECCV,} 2018, pp. 3-19. \href{https://arxiv.org/abs/1807.06521}{https://arxiv.org/abs/1807.06521}

\bibitem{b18} Roy A. G, Siddiqui S, Pölsterl S, et al, Squeeze \& excite guided few-shot segmentation of volumetric images, \emph{Medical Image Analysis,} 2020; 59: 101587. \href{https://arxiv.org/abs/1902.01314}{https://arxiv.org/abs/1902.01314}

\bibitem{b19} Cai Y. T and Wang Y, MA-UNet: An improved version of UNet based on multi-scale and attention mechanism for medical image segmentation, in \emph{Third International Conference on Electronics and Communication Network; and Computer Technology (ECNCT 2021),} SPIE, 2022; 12167: 205-211. \href{https://arxiv.org/abs/2012.10952}{https://arxiv.org/abs/2012.10952}

\bibitem{b20} Sinha A, and Dolz J, Multi-scale self-guided attention for medical image segmentation, \emph{IEEE Journal of Biomedical and Health Informatics}, 2021; 25(1): 121-130. \href{https://arxiv.org/abs/1906.02849}{https://arxiv.org/abs/1906.02849}

\bibitem{b21}  Chen L C, Zhu Y, Papandreou G, et al. Encoder-decoder with atrous separable convolution for semantic image segmentation, in \emph{Proceedings of the European conference on computer vision (ECCV),} 2018, pp. 801-818. \href{https://arxiv.org/abs/1802.02611}{https://arxiv.org/abs/1802.02611}

\bibitem{b22} Ronneberger O, Fischer P, and Brox T, U-Net: Convolutional networks for biomedical image segmentation, in \emph{International Conference on Medical Image Computing and Computer-Assisted Intervention,} Springer, 2015, pp. 234–241. \href{https://doi.org/10.48550/arXiv.1505.04597}{https://doi.org/10.48550/arXiv.1505.04597} 

\bibitem{b23} Çiçek Ö, Abdulkadir A, Lienkamp S S, et al. 3D U-Net: learning dense volumetric segmentation from sparse annotation, in \emph{Medical Image Computing and Computer-Assisted Intervention–MICCAI 2016: 19\textsuperscript{th} International Conference,} Athens, Greece, October 17-21, 2016, Proceedings, Part II 19: pp. 424-432. Springer International Publishing. \href{https://doi.org/10.48550/arXiv.1606.06650}{https://doi.org/10.48550/arXiv.1606.06650}

\bibitem{b24} Milletari F, Navab N, and Ahmadi S. A, V-Net: Fully convolutional neural networks for volumetric medical image segmentation, in \emph{2016 4\textsuperscript{th} International Conference on 3D Vision (3DV),} IEEE, 2016, pp. 565–571. \href{https://ieeexplore.ieee.org/abstract/document/7785132}{https://ieeexplore.ieee.org/abstract/document/ \\7785132}

\bibitem{b25} Xiao X, Lian S, Luo Z, et al, Weighted RES-Unet for high-quality retina vessel segmentation, in \emph{2018 9\textsuperscript{th} international conference on information technology in medicine and education (ITME)}, IEEE, 2018, pp. 327-331. \href{https://ieeexplore.ieee.org/abstract/document/8589312}{https://ieeexplore.ieee.org/abstract/document/\\8589312}

\bibitem{b26}  Li X, Chen H, Qi X, et al. H-DenseUNet: Hybrid densely connected UNet for liver and tumor segmentation from CT volumes, \emph{IEEE Transactions on Medical Imaging}, 2018; 37(12): 2663–2674. \href{https://ieeexplore.ieee.org/document/8379359}{https://ieeexplore.ieee.org/document/8379359}

\bibitem{b27} Ibtehaz N, and Rahman M. S, MultiResUNet: Rethinking the U-Net architecture for multimodal biomedical image segmentation. \emph{Neural networks,} 2020; 121: 74-87. \href{https://arxiv.org/abs/1902.04049}{https://arxiv.org/abs/1902.04049}

\bibitem{b28} Park D and Lee J. Hierarchical Attention Network for Few-Shot Object Detection via Meta-Contrastive Learning. \emph{arXiv preprint arXiv: 2208.07039,} 2022. \href{https://arxiv.org/abs/2208.07039v1}{https://arxiv.org/abs/2208.07039v1}

\bibitem{b29} Guo M, Lu C, Liu Z, et al. Visual attention network. Computational Visual Media. 2023; 9(4): 733-752. \href{https://arxiv.org/abs/2202.09741}{https://arxiv.org/abs/2202.09741}

\bibitem{b30} Kavur A. E, Gezer N. S, Barış M, et al, CHAOS challenge-combined (CT-MR) healthy abdominal organ segmentation, \emph{Medical Image Analysis,} 2021; 69: 101950. \href{https://arxiv.org/abs/2001.06535}{https://arxiv.org/abs/2001.06535}

\bibitem{b31} Zhuang X, Multivariate mixture model for cardiac segmentation from multi-sequence MRI, in \emph{International Conference on Medical Image Computing and Computer-Assisted Intervention,} Springer, 2016, pp. 581–588. \href{https://arxiv.org/pdf/1612.08820.pdf}{https://arxiv.org/pdf/1612.08820.pdf}

\bibitem{b32} Ding H, Sun C, Tang H, et al. Few-shot Medical Image Segmentation with Cycle-resemblance Attention, in \emph{Proceedings of the IEEE/CVF Winter Conference on Applications of Computer Vision.} 2023, pp. 2488-2497. \href{https://ieeexplore.ieee.org/document/10030099}{https://ieeexplore.ieee.org/document/10030099}


\end{thebibliography}
\end{document}